\documentclass[letterpaper]{article} % DO NOT CHANGE THIS
\usepackage{aaai23}  % DO NOT CHANGE THIS
\usepackage{times}  % DO NOT CHANGE THIS
\usepackage{helvet}  % DO NOT CHANGE THIS
\usepackage{courier}  % DO NOT CHANGE THIS
\usepackage[hyphens]{url}  % DO NOT CHANGE THIS
\usepackage{graphicx} % DO NOT CHANGE THIS
\usepackage{natbib}  % DO NOT CHANGE THIS AND DO NOT ADD ANY OPTIONS TO IT
\usepackage{caption} % DO NOT CHANGE THIS AND DO NOT ADD ANY OPTIONS TO IT
\usepackage{algorithm}
\usepackage{algorithmic}
\usepackage{newfloat}
\usepackage{listings}
\DeclareCaptionStyle{ruled}{labelfont=normalfont,labelsep=colon,strut=off} % DO NOT CHANGE THIS
\floatstyle{ruled}
\newfloat{listing}{tb}{lst}{}
\floatname{listing}{Listing}
\usepackage{graphicx}
\usepackage{algorithm}
\usepackage{algorithmic}
\usepackage{amsmath,amssymb}
\usepackage{amsfonts}
\usepackage{hyperref}
\usepackage{multirow}
\usepackage{makecell}
\usepackage{booktabs}
\usepackage{enumitem}

\usepackage{stfloats}
\usepackage{color}
\usepackage[table]{xcolor}
\usepackage{colortbl}
\usepackage {float}
\usepackage{booktabs}

\newcommand{\rc}[1]{\textcolor{red}{#1}}
\newcommand{\tb}[1]{\textbf{#1}}
\newcommand{\D}{\Delta}
\newcommand{\x}{\boldsymbol{\ast}}
\newcommand{\ttt}[1]{\texttt{#1}}

\title{A Neural Span-Based Continual Named Entity Recognition Model}
\author {
    Yunan Zhang\textsuperscript{\rm 1},
    Qingcai Chen\textsuperscript{\rm 1, \rm 2}\footnote{Corresponding author}
}
\affiliations {
    \textsuperscript{\rm 1} Harbin Institute of Technology (Shenzhen), Shenzhen, China \\
    \textsuperscript{\rm 2} Peng Cheng Laboratory, Shenzhen, China \\
    yunanzhang0@outlook.com, qingcai.chen@hit.edu.cn
}
\begin{document}

\maketitle

\begin{abstract}
Named Entity Recognition (NER) models capable of Continual Learning (CL) are realistically valuable in areas where entity types continuously increase (e.g., personal assistants).
Meanwhile the learning paradigm of NER advances to new patterns such as the span-based methods.
However, its potential to CL has not been fully explored.
In this paper, we propose \tb{SpanKL}, a simple yet effective \tb{Span}-based model with \tb{K}nowledge distillation (KD) to preserve memories and multi-\tb{L}abel prediction to prevent conflicts in CL-NER.
Unlike prior sequence labeling approaches, the inherently independent modeling in span and entity level with the designed coherent optimization on SpanKL promotes its learning at each incremental step and mitigates the forgetting.
Experiments on synthetic CL datasets derived from OntoNotes and Few-NERD show that SpanKL significantly outperforms previous SoTA in many aspects, and obtains the smallest gap from CL to the upper bound revealing its high practiced value. The code is available at \url{https://github.com/Qznan/SpanKL}.
\end{abstract}

\section{Introduction}
Deep neural models have demonstrated impressive performances on standard tasks, but their abilities to continually learn a sequence of tasks still remain a real challenge as the requirement to learn to adapt to new information and meanwhile to retain prior acquired knowledge.
It suffers from the well-known catastrophic forgetting or interference issue particularly with the advent of new tasks from changing distribution. This also troubles the studies on Continual Learning Named Entity Recognition (CL-NER) task \cite{biesialska-etal-2020-continual}.

Recent explorations on CL-NER with promising results \cite{Monaikul2021ContinualLF,xia-etal-2022-learn} formulate the problem in a class-incremental setting, of which the most prominent feature is that each task only contains the annotations of entity types defined to be learned in that task.
In this context, CL-NER in NLP is more akin to continual object detection than image classification in Computer Vision.
Because previously learned entity type's mentions may appear in the samples trained in the current task but without the relevant annotations.
These false negative labels will unavoidably compel models to forget old knowledge to fit the new conflicting one.
To address this backward incompatibility, they leverage knowledge distillation to predict the distilled (pseudo) labels using the previously learned model (teacher) on current samples, and then learn a current model (student) jointly by these labels and the current golden labels.

\begin{figure}
\centering
\includegraphics[width=2.5in]{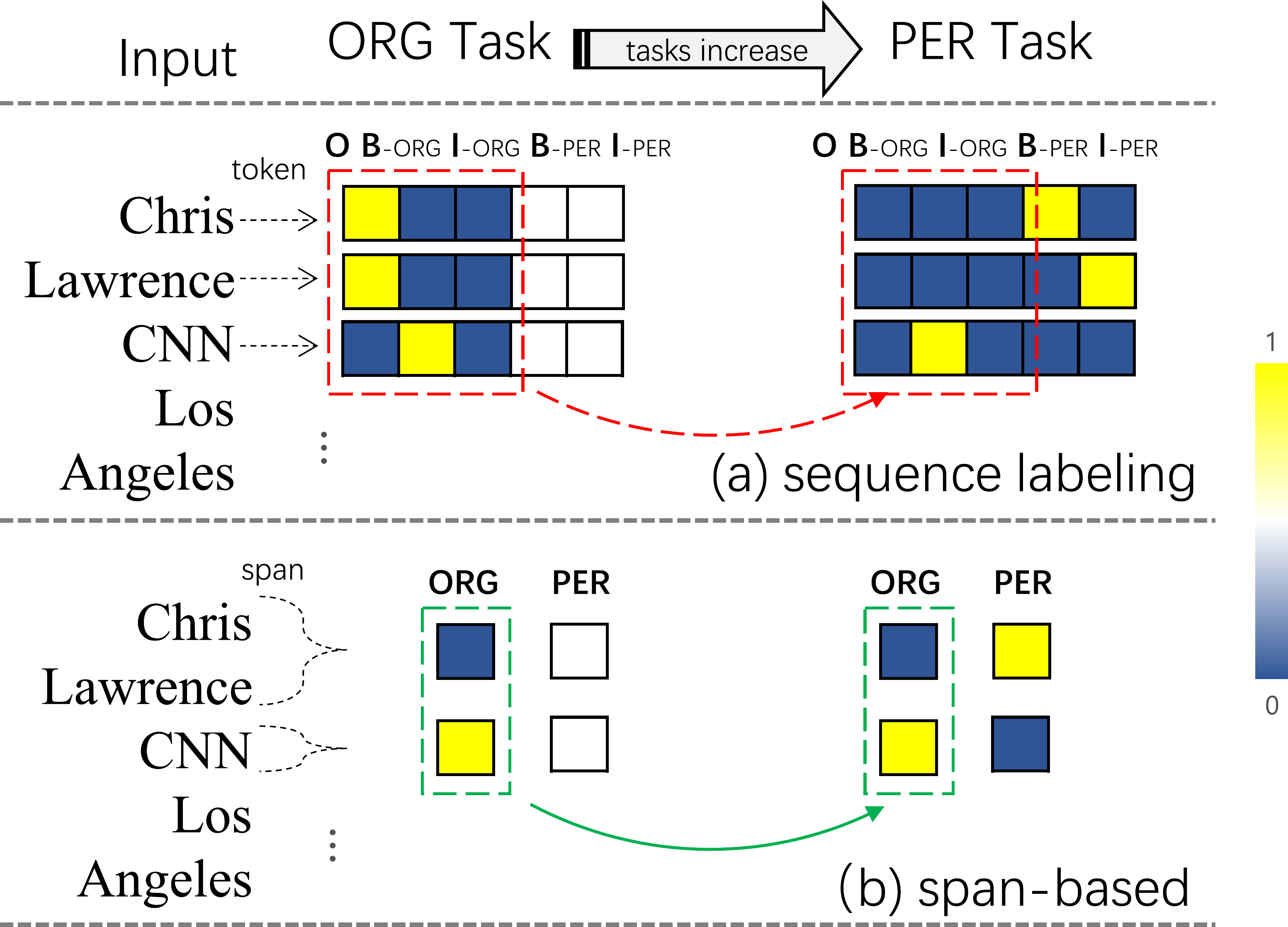}
\caption{The incoherent optimization across CL tasks of sequence labeling methods (\textcolor{red}{dash arrow}) compared with SpanKL (\textcolor{teal!50!green}{solid arrow}) at the learning targets.}
\label{motivation}
\end{figure}

However, an opposite but easily neglected case is that the non-entity mentions learning currently may belong to a certain entity type to be learned in future tasks, and we find it forward incompatible for the traditional sequence labeling methods to successively handle them in CL. 
Specifically, for mentions whose entity types will be learned in future tasks but not in current, sequence labeling methods will assign a global \ttt{O} tag indicating non-(any)-entity in the current task.
But then they need to frequently change the learning target of these mentions when the future relevant tasks arrive.
As shown in Fig.~\ref{motivation}a, the model's output logit vector should be altered, e.g., from predicting \ttt{O} to predicting \ttt{B-PER} for the first token of mention \textit{Chris Lawrence} when PER task comes after the ORG task.
This incoherent optimization will force the model to frequently update the previously learned parameters, thus we consider aggravates the catastrophic forgetting or interference.

A natural solution for this is regarding any non-entity mentions in the current task only as non-(certain)-entity, and if possible, in a more efficient span-level, i.e., the span of mention \textit{Chris Lawrence} is \ttt{O-ORG} rather than \ttt{O} in the ORG task (Fig.~\ref{motivation}b).
It conforms to the Risk Minimization principle for safely taking one mention as non-ORG in ORG task, and thus without conflict further taking it as PER in the future PER task, which essentially converts the entity recognition into a binary classification problem.

Motivated by this, we propose SpanKL, a better CL-adapted architecture for NER.
As a span-based model, it enumerates all spans and learns their representations to classify entities.
For CL, it is equipped with the multi-label learning seamlessly at the span classification, and with the knowledge distillation in span and entity-level on the yielding Bernoulli distribution.
This gives the following advantages towards coherently optimizing:
1) It's backward compatible yet in CL as freely fusing the KD technique to preserve old knowledge.
2) It's also forward compatible in CL as identifying each entity type by binary classifying to reduce interference in future tasks.
3) The independent modeling in span and entity-level has better learning and distillation ability and is flexible for sequentially learning various entity types.
4) Benefiting from the span-based method any case of nested entities\footnote{Overlapped spans are same or different entities, single span has different entities, entity types in one or across multiple tasks.} are supported in CL-NER.

In evaluation, compared with synthesizing CL dataset from OntoNotes restricted to learning only single entity type per task recently, we newly use the elaborate Few-NERD with adequate entity types that enable at least 6 types in each of the 8 synthetic tasks for a comprehensive study. We also detect 4 synthetic setups existing in recent work influential to the performance, and thus report results in all these setups for a fair comparison.

Results on both datasets empirical show that SpanKL significantly outperforms the existing CL-NER baselines and achieves new SoTA. 
The performance gaps between CL and the upper bound non-CL also suggest that SpanKL can almost eliminate forgetting on the relatively simple OntoNotes, reaching a new level in practice.  

Our contributions are: (i) We build a simple span-based architecture to achieve coherent optimization in CL-NER, which can serve as a immediate and strong baseline empirically. (ii) We align the comparison in different synthetic setups for recent models, and explore the more realistic CL-NER scenario by a new benchmark for future research.

\section{Related Work}
\textbf{Named Entity Recognition} is a fundamental task in information extraction to recognize the predefined entity types from text \cite{lample-2016-lstmcrfner}. 
Recently, the learning paradigm shift of NER models such as converting the traditional sequence labeling-based manner into the span-based is widely explored and obtains promising results \cite{yu-2020-biaffine:span-based,fu-2021-spanner,Xu-2021-SMHSA}.
Instead of tagging each token under the elaborate tagging scheme, e.g., IOB, span-based methods directly enumerate all possible spans and classify them into predefined entity type or non-entity.
It provides more granular modeling in each span and supports the case of nested entities.

\noindent \textbf{Continual Learning} solves the problem where the training signals arrive as a stream.
Commonly these training data may be sampled from a progressively changing distribution (i.e., non-stationary distribution) or simply belong to a sequence of different tasks \cite{chen2018lifelong}.
Compared to the standard full data training, CL models are vulnerable to catastrophic forgetting (CF), especially when the incoming new tasks have different data distributions from the old tasks \cite{goodfellow2013empirical}.
Early studies mainly focus on the task-incremental learning regime, which relies on explicit task-IDs during inference \cite{van2019three}.
By contrast, class-incremental learning is more challenging since it needs to concurrently distinguish between all classes from all tasks learned so far, incurring confusion \cite{masana2020class}.

Generally three veins exist to resolve the CF issue:
1) Regularization-based methods constrain the model weights updating to maintain the performance of old tasks or impose sparsity on the weights in order to activate a subset of neurons regarding different tasks \cite{kirkpatrick2017overcoming,serra2018overcoming}. This can be also archived by KD \cite{li2017learning}.
2) Rehearsal-based methods typically reserve a small set of samples from old tasks in memory for jointly training in new tasks, which mimic the \textit{i.i.d} protocol. These replayed samples also can be generated by a generative model \cite{rebuffi2017icarl,castro2018end,shin2017continual}.
3) Isolation-based methods explicitly allocate different parameters to each task by dynamically growing architecture \cite{xu2018reinforced}.

\noindent \textbf{Continual Learning NER}.
Prior CL-related works mostly focused on Computer Vision, but recent explorations dedicated to NLP typically NER have emerged.
\citeauthor{chen2019transfer} \shortcite{chen2019transfer} first study the knowledge transfer of the sequence labeling NER model from the source domain to target domain with new entities.
They use a neural adapter module for diverse distributions of entities between tasks.

In a more formal CL protocol, recently AddNER, ExtendNER~\cite{Monaikul2021ContinualLF} and L\&R~\cite{xia-etal-2022-learn} are designed to solve CL-NER first under a class-incremental setting, where data of each task are only annotated by the currently learning entities practically.
They all base on the sequence labeling methods with knowledge distillation.
Categorized by the layout of the model output \cite{de2021continual},
AddNER is multi-head as each task uses an individual head, while ExtendNER is single-head as all tasks share a unified head.
The main difference is AddNER provides distinct \ttt{O} tags for each head whereas ExtendNER only preserves a global \ttt{O} tag.
L\&R basically equips ExtendNER with a reviewing stage to generate synthetic samples by a language model, to prevent deterioration of distillation in case samples learning currently lack the old entity mentions.

Although AddNER uses multiple \ttt{O} tags, each one is specific to a task instead of to an entity type as in SpanKL.
It will still face the forward incompatibility issue existing in ExtendNER typically when tasks with multiple entity types arrive.
Moreover, AddNER need to design a heuristic strategy to combine all heads outputs for valid prediction, and ExtendNER need to manually pad small constants to distilled labels for alignment.
Their cooperations with KD are also cumbersome with the frequent switch on the learning labels (distilled or golden) at each token.
Without these defects, SpanKL, to our best knowledge, is the first study on the potential of span-based model to solve CL-NER.

\section{Method}
\subsection{Problem Formulation}
We follow the recent works to formulate CL-NER under class-incremental setting \cite{Monaikul2021ContinualLF,xia-etal-2022-learn}.
Given a sequence of tasks $T_1,T_2,...T_l$ and the corresponding sets of entity types $E_1,E_2,...E_l$ defined to be continually learned, the $l$-th task has its own training set $D_l$ only annotated for the the entity types
$E_l=\{e_{l1},e_{l2},...\}$.
Entity types in different tasks are non-overlapping, e.g., if ORG is learned in $T_1$ then it will not be learned in other tasks.
But mentions of diverse entity types are allowed to overlap whether learned from one or different tasks, i.e., no restriction to any case of nested entities.

At the first step ($l=1$), we train model $M_1$ on $D_1$ from scratch to recognize entities of types $E_1$.
At the following $l$-th incremental step ($l>1$), we train $M_l$ on $D_l$ based on the previous learned model $M_{l-1}$ to recognize entities of types learned so far $\cup_{i=1}^{l}E_i$.

\subsection{SpanKL NER Model}
\begin{figure}[htbp]
\centering
\includegraphics[width=2.5in]{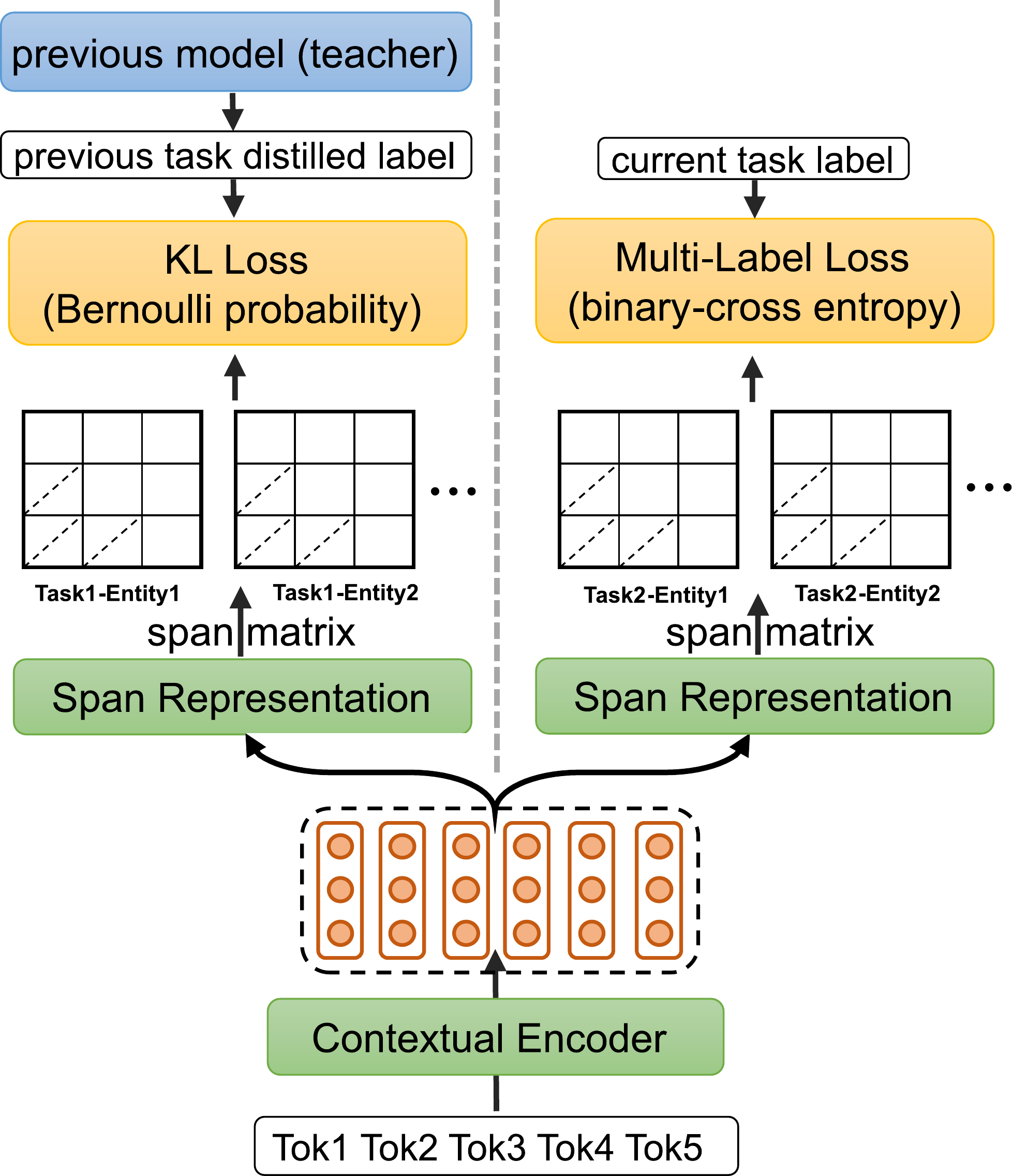}
\caption{Overall architecture of SpanKL including a shared contextual encoder for all tasks and the distinct span representation layer for the entity types in each task.
Bernoulli KL loss and BCE loss are computed for previously learned and currently learning entities, respectively, on the corresponding entity-related span matrix.
}
\label{model}
\end{figure}

We introduce the simple yet effective SpanKL (see Fig.~\ref{model}) to sequentially learn each task.
Given an input sentence $X$ with $n$ tokens $[x_1, x_2, ..., x_n]$, we define $s_{ij}$ as span comprising continuous tokens that starts with $x_i$ and ends with $x_j$, $1\!\leqslant\!i\leqslant\!j\leqslant\!n$.
Assume there are $K$ entity types to be learned at the $l$-th incremental step, the goal of SpanKL is to represent each span into $\mathbf{h}^{s_{ij}}=[h^{s_{ij}}_1, h^{s_{ij}}_2,...h^{s_{ij}}_k]$ by span modeling and to perform binary classification of these $K$ entity types.
It consists of the contextual encoder, span representation layer and multi-label loss layer with the knowledge distillation, as described below.

\noindent \textbf{Contextual Encoder} captures the dependence between tokens within input sentences and can be implemented with the widely-used CNN, RNN or PLM models.
We use $\mathbf{E}=[\mathbf{e}_1,\mathbf{e}_2,...,\mathbf{e}_n]\!\in\!\mathbb{R}^{n \times d^e}$ to denote the embedding vectors of input $X$ after embedding, then feed it into the contextual encoder to get the contextualized hidden vectors $\mathbf{H} = [\mathbf{h}_1,\mathbf{h}_2,...,\mathbf{h}_n]\!\in\!\mathbb{R}^{n \times d^h}$ for each token as:
\begin{gather}
\mathbf{E} = \text{Embed}(X), \mathbf{H} = \text{CtxEnc}(\mathbf{E}),
\end{gather}
where \textit{Embed} is embedding layer, \textit{CtxEnc} is contextual encoder. $d_e$ and $d_h$ is dimension of embedding and hidden, respectively. The encoder is shared for all tasks.

\noindent \textbf{Span Representation Layer} performs span modeling as:
\begin{gather}
\mathbf{h}^{s_{ij}}\!=\!\text{SpanRep}(\mathbf{h}_i, \mathbf{h}_{i+1}, ..., \mathbf{h}_{j}),
\end{gather}
where the span representation is generated from the related token representations and various design have been fully explored. As the boundary tokens of entities are most informative, \citeauthor{yu-2020-biaffine:span-based} \shortcite{yu-2020-biaffine:span-based} models both the start and end feature space of tokens constituting the spans with the biaffine interaction. \citeauthor{Xu-2021-SMHSA} \shortcite{Xu-2021-SMHSA} further models the entity types feature space as the heads in multi-head (additive) attention and \citeauthor{kexuefm-8373} \shortcite{kexuefm-8373} uses multi-head (dot-product) attention\footnote{Essentially a decomposed non-bias biaffine (i.e., bilinear) as $\mathbf{h}_i^\intercal \mathbf{W}^{\mathfrak{s}\intercal} \mathbf{W}^\mathfrak{e}\mathbf{h}_j\Leftrightarrow \mathbf{h}_i^\intercal \mathbf{W} \mathbf{h}_j$.}. We adopt the latter simple manner with the complete weights separation in these three feature space. Specifically, for each entity type two distinct single-layer feed-forward networks (FFN) regarding the start and end modeling (totaled $2K$ distinct FNNs) are used before the scaled dot-product interaction as:
\begin{gather}
h^{s_{ij}}_k=\text{FFN}^{\mathfrak{s},k}(\mathbf{h}_i)^\intercal \text{FFN}^{\mathfrak{e},k}(\mathbf{h}_j) \times (d^{o})^{-0.5}, \label{compute_startend}
\end{gather}
where $\mathfrak{s}$, $\mathfrak{e}$, $k$ denote the start, end, k-th entity type. $d^{o}$ is output dimension of all the $2K$ FNNs.
We believe this clearly separated modeling w.r.t each entity type at each span (i.e., in span and entity-level) can facilitate the learning and distillation and alleviate the interference between multiple tasks. As tasks increase, we simply add more span representation layers (essentially the inner FFNs) dedicated to the new task.

\textit{Span Matrix} is conceptually introduced for better description (Fig.~\ref{model}). We organized all $h^{s_{ij}}_k$ related to the $k$-th entity into the upper triangle region of matrix $\mathcal{M}^k\!\in\!\mathbb{R}^{n\times n},\mathcal{M}^k_{ij}\!=\!h^{s_{ij}}_k$, where the row and column indicate the start and end. 

\noindent \textbf{Multi-Label Loss Layer}. To ensure the forward comparability expectantly, we formulate the final span classification as multi-label prediction. Specifically, we compute the Binary Cross Entropy (BCE) loss after \textit{sigmoid} activation on the predicted logit in the span matrix with the golden label.
Compared with the popular multi-class manner, i.e., the Cross Entropy (CE) loss with \textit{softmax} activation, it disentangles different entity types when normalizing the logit into the probability over whether a single or multiple tasks.
Each entity type is independently binary classified and the BCE loss is computed as:
\begin{gather}
\hat{p}(k|s_{ij}) = \text{sigmoid}(h^{s_{ij}}_k), \\
\mathcal{L}_{BCE} = -\sum_{i=1}^n\sum_{j=1}^n\sum_{k=1}^Lp(k|s_{ij})\log(\hat{p}(k|s_{ij}) \nonumber \\
 + (1-p(k|s_{ij}))\log(1-\hat{p}(k|s_{ij})),
\end{gather}
where $p(k|s_{ij})$ is the golden label, $\hat{p}(k|s_{ij})$ is the predicted label.
$\mathcal{L}_{BCE}$ is computed only upon the current entity types' span matrices.

\noindent \textbf{Knowledge Distillation}.
To ensure the backward comparability yet, we use KD \cite{hinton2015distilling,DBLP:conf/icml/GuptaCRS20} to prevent forgetting old entities.
At the $l$-th incremental step ($l>1$), we first make a one-off prediction using the previously learned model $M_{l-1}$ (teacher) on the whole current training set $D_l$ for the entity types learned up to previous step $\cup_{i=1}^{l-1}E_i$.
This yields the Bernoulli distribution as the soft distilled label $\tilde{p}$ of every span for every old entity type.
These pseudo labels are used to compute the Bernoulli KL divergence loss with the current model $M_{l}$ (student) as:
\begin{gather}
\mathcal{L}_{KD} = \sum_{i=1}^n\sum_{j=1}^n\sum_{k=1}^L
\tilde{p}(k|s_{ij})(\log(\tilde{p}(k|s_{ij})\!-\!\log(\hat{p}(k|s_{ij})) \nonumber \\
+ (1\!-\!\tilde{p}(k|s_{ij}))(\log(1\!-\!\tilde{p}(k|s_{ij}))\!-\!\log(1\!-\!\hat{p}(k|s_{ij}))),
\end{gather}
where $\tilde{p}(k|s_{ij})$ is the soft distilled label to be imitated.
$\mathcal{L}_{KD}$ is computed only upon the span matrices of old entity types.

The final loss used in the multiple epochs training after the one-off prediction at each step is the weighted sum as:
\begin{gather}
\mathcal{L} =  \alpha \mathcal{L}_{BCE} + \beta \mathcal{L}_{KD},
\end{gather}
where $\alpha$ and $\beta$ are the weights of both losses.

\section{Experiments}
\subsection{Datasets}
We follow recent works \cite{Monaikul2021ContinualLF,xia-etal-2022-learn} to convert the widely used standard NER corpora into separated datasets acting as a series of CL synthetic tasks in class-incremental setting.
Besides their usage of OnteNotes \cite{Pradhan13ontonote} with only a single entity type per task, we further use the larger and more complicated Few-NERD \cite{ding2021few} that enables multiple entity types per task.

\noindent \textbf{OntoNotes-5.0 English}\footnote{\url{https://catalog.ldc.upenn.edu/LDC2013T19}} is annotated for 18 entity types, we follow the recent works to select the following types to ensure sufficient samples for training: \textit{Organization (ORG), Person (PER), Geo-Political Entity (GPE), Date (DATE), Cardinal (CARD), Nationalities and Religious Political Group (NORP)}. Each type is assigned to a synthetic CL task.

\noindent \textbf{Few-NERD (SUP)}\footnote{\url{https://ningding97.github.io/fewnerd}} is hierarchically annotated for 8 coarse-grained and 66 fine-grained entity types. It's proposed for few-shot research but we adopt the normal supervised full version. We construct each task via each coarse-grained types and thus each task contains its related multiple fine-grained entity types that will be evaluated. This is more practical since each task is a domain with multiple relative entity types. The coarse-grained types include \textit{Location (LOC), Person (PER), Organization (ORG), Other (OTH), Product (PROD), Building (BUID), Art (ART), Event (EVET)}
and the related fine-grained types are shown in Appendix.
% \footnote{Appendix is available at \url{https://github.com/Qznan/SpanKL}.}.

\subsection{Synthetic Setup}
Beyond the model architecture, an important but easily neglected detail that we consider largely affects the performance is how to divide the original dataset to construct the synthetic CL dataset.
However, we find the diversity of this synthetic setup in recent works thus making their comparisons unreliable.

The synthetic setup includes two aspects: 1) To separate the original training/dev set into a series of CL tasks,
\citeauthor{Monaikul2021ContinualLF} \shortcite{Monaikul2021ContinualLF} commonly divides samples randomly into disjoint tasks, while \citeauthor{xia-etal-2022-learn} \shortcite{xia-etal-2022-learn} typically filter samples having the entity types defined to be learn in that task to compose its datasets, which we refer to as \textit{Split} and \textit{Filter}, respectively.
2) To form the test set evaluated in the series of CL tasks, 
\citeauthor{Monaikul2021ContinualLF} \shortcite{Monaikul2021ContinualLF} maintain full of samples in original test set,
while \citeauthor{xia-etal-2022-learn} \shortcite{xia-etal-2022-learn}, again, filter samples having the entity types learned so far as test set,
which we refer to as \textit{All} and \textit{Filter}, respectively.

There hence exist 4 combinations of synthetic setup from the two aspects above: Split-All, Split-Filter, Filter-All, Filter-Filter. 
In the training, compared to the disjoint Split-$\x$ setups, Filter-$\x$ enable repetitive learning of the sample with multiple entity types assigned to different tasks, but lack the learning of samples without any entity mentions (namely non-entity sample).
During testing, $\x$-All setups are more challenging than $\x$-Filter with non-entity samples that require the model's denial ability.
Due to the above influences, we test all of them for comprehensive evaluation.

After synthesizing, the training/dev set of each task is only allowed to contain the task-predefined single/multiple entity type(s) for OntoNotes/Few-NERD, which means we will replace the irrelevant entity types with non-entity, e.g., erasing the annotations of ORG on samples assigned to task learning PER. Similarly, the test set of each task is only allowed to contain the entity types learned up to that task.

\begin{table*}[htbp]\small
\centering
\tabcolsep=5.3pt
\begin{tabular}{@{}lr|llllll|llllll@{}}
\toprule

                                                                         &                  & \multicolumn{6}{c|}{Train/Dev(\tb{Split})}                                       & \multicolumn{6}{c}{Train/Dev(\tb{Filter})}                             \\
                                                                         &                  &     Step1  &      Step2 &      Step3 &      Step4 &      Step5 &      Step6 &      Step1 &      Step2 &      Step3 &      Step4 &      Step5 &      Step6 \\ \midrule 
                                                                         & \rc{non-CL}      &     82.52 & 86.42 & 87.32 & 88.84 & 89.62 & 89.27                           & 69.22 & 79.84 & 83.86 & 83.22 & 85.22 & 85.14                               \\ 
                                                                         & AddNER$^\star$   &     82.52 & 83.90 & 84.66 & 85.02 & 85.48 & 85.03                           &      69.22 & 76.72 & 78.22 & 78.29 & 79.44 & 79.03                          \\ \rowcolor{gray!20}
                                                                         &$\D$              &     -0.00 & -2.52 & -2.66 & -3.82 & -4.14 & -4.24                           &      -0.00 & -3.12 & -5.64 & -4.93 & -5.78 & -6.11                         \\ \cmidrule(l){2-14} 
                                                                         
                                                                         & \rc{non-CL}      & 82.79 & 86.49 & 87.70 & 88.46 & 89.02 & 89.19                               &      69.90 & 77.76 & 81.50 & 81.60 & 83.16 & 81.28                           \\ 
                                                                         & ExtendNER$^\star$&     82.79 & 83.54 & 84.48 & 84.67 & 85.12 & 84.96                          &      69.90 & 74.14 & 73.72 & 72.88 & 72.58 & 69.29                          \\ \rowcolor{gray!20}
                                                                         &$\D$              &     -0.00 & -2.95 & -3.22 & -3.79 & -3.90 & -4.23                            &      -0.00 & -3.62 & -7.78 & -8.72 & -10.58 & -11.99                         \\  \cmidrule(l){2-14}
                                                                         
                                                                         & \rc{non-CL}      &    85.60 & 88.16 & 88.64 & 89.39 & 89.69 & 89.74                            &      72.78 & 79.60 & 83.48 & 84.28 & 87.46 & 86.48                               \\ 
\multirow{-9}{*}{\cellcolor{white}\makecell[c]{Test\\(\tb{All})}}        &\tb{SpanKL}   &     85.60 & 87.92 & 88.22 & 88.76 & 89.02 & \tb{88.98}                         &      72.78 & 79.46 & 81.89 & 81.96 & 81.81 & \tb{79.31}                         \\ \rowcolor{gray!20} 
                                                                         &$\D$              &     -0.00 & -0.24 & -0.42 & -0.63 & -0.67 & \tb{-0.76}                      &      -0.00 & -0.14 & -1.59 & -2.32 & -5.65 & \tb{-7.17}                           \\ \midrule 
                                                                         
                                                                         & \rc{non-CL}      &     84.74 & 87.74 & 88.44 & 89.84 & 90.35 & 90.00                          &       90.78 & 91.54 & 90.76 & 90.60 & 90.50 & 90.48                               \\ 
                                                                         & AddNER$^\star$   &     84.74 & 85.44 & 85.73 & 86.00 & 86.28 & 85.98                           &      90.78 & 89.82 & 88.92 & 87.20 & 86.16 & 85.82                         \\ \rowcolor{gray!20} 
                                                                         &$\D$              &     -0.00 & -2.30 & -2.71 & -3.84 & -4.07 & -4.02                           &      -0.00 & -1.72 & -1.84 & -2.40 & -3.34 & -4.66                         \\ \cmidrule(l){2-14} 
                                                                         
                                                                         & \rc{non-CL}      & 84.81 & 87.86 & 88.73 & 89.36 & 89.74 & 89.88                               &      90.62 & 91.70 & 91.02 & 90.79 & 90.92 & 90.10                               \\ 
                                                                         & ExtendNER$^\star$&     84.81 & 85.10 & 85.76 & 85.83 & 86.07 & 86.00                           &      90.62 & 88.92 & 87.55 & 86.30 & 84.77 & 81.37                          \\\rowcolor{gray!20}
                                                                         &$\D$              &     -0.00 & -2.76 & -2.97 & -3.53 & -3.67 & -3.88                            &      -0.00 & -2.78 & -3.47 & -4.49 & -6.15 & -8.73                          \\ 
                                                                         
                                                                         & L\&R$^\circ$     &     -      &      -     &      -     &      -     &      -     &      -     &      92.06 &      88.09 &      85.69 &      83.79 &      83.38 &      83.02 \\ \cmidrule(l){2-14}  

                                                                         & \rc{non-CL}      & 87.81 & 89.58 & 89.97 & 90.48 & 90.34 & 90.43                               & 92.37 & 92.65 & 92.78 & 92.06 & 92.10 & 91.90                               \\ 
\multirow{-10}{*}{\makecell[c]{Test\\(\tb{Filter})}}                     &\tb{SpanKL}   &     87.81 & 89.28 & 89.46 & 89.74 & 89.80 & \tb{89.78}                          &      92.37 & 90.81 & 90.38 & 89.50 & 89.18 & \tb{88.07}                       \\ \rowcolor{gray!20}
                                                                         &$\D$              &     -0.00 & -0.30 & -0.51 & -0.74 & -0.54 & \tb{-0.65}                      &      -0.00 & -1.84 & -2.40 & -2.56 & -2.92 & \tb{-3.83}                          \\
                    
\bottomrule
\end{tabular}
\caption{Macro-F1 of different models at each step in four synthetic setups on OntoNotes. The gap ($\D$=CL$-$non-CL) is shaded. $\star$ is our reimplementation, $\circ$ is result referred to the original paper.
Highest value at the final step in CL is bolded with its gap.}
\label{performace_onto}
\end{table*}

\subsection{Metrics}
Given a certain task at each step, we train the models on its training set and report the performance of the following metrics on its test set relying on the best performance of its dev set.
We follow the existing 6 permutations of tasks on OntoNotes, and randomly sample 4 permutations on Few-NERD to factor out the influence of the learning order
(see Appendix).
Results are averaged over all permutations unless otherwise specified.

\noindent \tb{Macro-average F1.}
For OntoNotes, we follow \cite{Monaikul2021ContinualLF,xia-etal-2022-learn} to compute the F1 score for each entity type and report the macro-average F1 score over all types learned so far at each step.
For Few-NERD, we compute the F1 score for each fine-grained type. Since the fine-grained types under the same coarse-grained type are unbalanced, we compute the micro-average F1 score for the aggregated coarse-grained type but still, fairly report the macro-average F1 score over all coarse-grained types learned so far at each step.

\noindent \tb{Gap between CL and non-CL}. For each model, we also individually evaluate a non-CL complete setting \cite{Monaikul2021ContinualLF,xia-etal-2022-learn} at each step as the upper bound of CL, which means we add all datasets of previous tasks to the current with the fully-annotated entity types defined to be learned so far during current task's training.
We report the gap between CL and non-CL involving each model to fairly consider their differential learning capacities.

\subsection{Baselines \& Implementation Details}
We compare SpanKL with the following baselines \cite{Monaikul2021ContinualLF,xia-etal-2022-learn}: 
single head \textbf{AddNER}, multi-head \textbf{ExtendNER} and its enhanced version \textbf{L\&R}.
Notably, these sequence labeling models do not adopt usual CRF possibly to avoid the intractable structural KD \cite{wang-etal-2021-structural},
thereby the tag transition risk may increase with more entity types introduced typically in single-head ExtendNER. This interests us in verifying whether multi-head AddNER is relatively better via more tests.
We reimplement AddNER, ExtendNER, and refer to the results from the original paper for L\&R.
Note that AddNER and ExtendNER are only evaluated in Split-All setup in the original paper, while the current SoTA L\&R is only evaluated in Filter-Filter.

We use \textit{bert-large-cased} from HuggingFace \cite{wolf2019huggingface} as the contextual encoder for all models ($d^h\!=\!1024$) followed by a 0.1 dropout.
we set $d^{o}\!=\!50$ for all subsequent FFNs in SpanKL. We set $\alpha\!=\!\beta\!=\!1$ for all models.
All parameters are fine-tuned by AdamW optimizer \cite{loshchilov2017decoupled}, with learning rate (lr) $5e^{-5}$ and $1e^{-3}$ for bert encoder and the rest networks.
The lr is scheduled by warmup at first 200 steps followed by a cosine decay.
We limit to 512 max length of sentence after the widely used BPE tokenization in PLMs and only use the representation

\begin{table}[H]\footnotesize
\tabcolsep=1.5pt
\begin{tabular}{@{}r|llllllll@{}}
\toprule
                      & \multicolumn{8}{c}{Train/Dev(\tb{Split})}                                       \\
                      &     Step1  &      Step2 &      Step3 &      Step4 &      Step5 &      Step6 &      Step7 &      Step8 \\ \midrule
\multicolumn{9}{l}{Test(\tb{All})} \\
\rc{non-CL}           &     64.01 & 62.25 & 61.88 & 61.07 & 61.28 & 62.43 & 63.83 & 63.65 \\
AddNER$^\star$        &     64.01 & 61.32 & 60.54 & 59.43 & 58.74 & 59.32 & 60.41 & 59.32                                     \\ \rowcolor{gray!20}
$\D$                  &     -0.00 & -0.93 & -1.34 & -1.64 & -2.54 & -3.11 & -3.42 & -4.33 \\ \cmidrule(l){1-9} 

\rc{non-CL}           &     64.06 & 62.08 & 61.76 & 60.94 & 61.29 & 62.02 & 63.62 & 63.32 \\
ExtendNER$^\star$     &     64.06 & 59.02 & 57.05 & 55.72 & 55.46 & 55.96 & 56.85 & 56.16 \\ \rowcolor{gray!20}
$\D$                  &     -0.00 & -3.06 & -4.71 & -5.22 & -5.83 & -6.06 & -6.77 & -7.16 \\ \cmidrule(l){1-9}

\rc{non-CL}           &     67.81 & 65.22 & 64.97 & 64.18 & 64.22 & 64.94 & 66.10 & 65.76 \\
\tb{SpanKL}       &     67.81 & 64.16 & 63.62 & 62.31 & 61.67 & 62.17 & 63.24 & \tb{62.15} \\ \rowcolor{gray!20}
$\D$                  &     -0.00 & -1.06 & -1.35 & -1.86 & -2.55 & -2.77 & -2.86 & \tb{-3.61} \\ \midrule
\multicolumn{9}{l}{Test(\tb{Filter})} \\

\rc{non-CL}           &     66.97 & 64.22 & 63.28 & 62.32 & 62.44 & 63.16 & 64.30 & 64.04 \\
AddNER$^\star$        &     66.97 & 63.08 & 61.80 & 60.62 & 59.86 & 60.07 & 60.92 & 59.74                                    \\ \rowcolor{gray!20}
$\D$                  &     -0.00 & -1.14 & -1.48 & -1.70 & -2.58 & -3.09 & -3.38 & -4.30    \\ \cmidrule(l){1-9}

\rc{non-CL}           &     66.89 & 64.05 & 63.22 & 62.31 & 62.48 & 62.79 & 64.10 & 63.70 \\
ExtendNER$^\star$     &     66.89 & 61.66 & 59.04 & 57.52 & 57.16 & 57.07 & 57.58 & 56.75 \\ \rowcolor{gray!20}
$\D$                  &     -0.00 & -2.39 & -4.18 & -4.79 & -5.32 & -5.72 & -6.52 & -6.95 \\ \cmidrule(l){1-9}

\rc{non-CL}           &     70.59 & 67.06 & 66.24 & 65.35 & 65.34 & 65.66 & 66.58 & 66.13 \\
\tb{SpanKL}       &     70.59 & 65.80 & 64.75 & 63.36 & 62.69 & 62.87 & 63.69 & \tb{62.50} \\ \rowcolor{gray!20}
$\D$                  &     -0.00 & -1.26 & -1.48 & -1.99 & -2.66 & -2.79 & -2.90 & \tb{-3.63} \\

\bottomrule
\end{tabular}
\caption{Macro-F1 on Few-NERD in Split-$\x$ setups.}
\label{performace_few}
\end{table}

\noindent of the first piece of word to represent this word after bert contextual encoder.
For the SpanKL output, we aggregate the predicted overlapped entities into a flat one (only keep the entity with the highest predicted score and discard the overlapping others).
We do not over-tune hyperparameters between corpora, except setting batch size 32, 24 and maximum epoch 10, 5 on OntoNotes, Few-NERD, respectively, to train on a V100 GPU.

\begin{figure*}[htbp]
\centering
\includegraphics[width=6in]{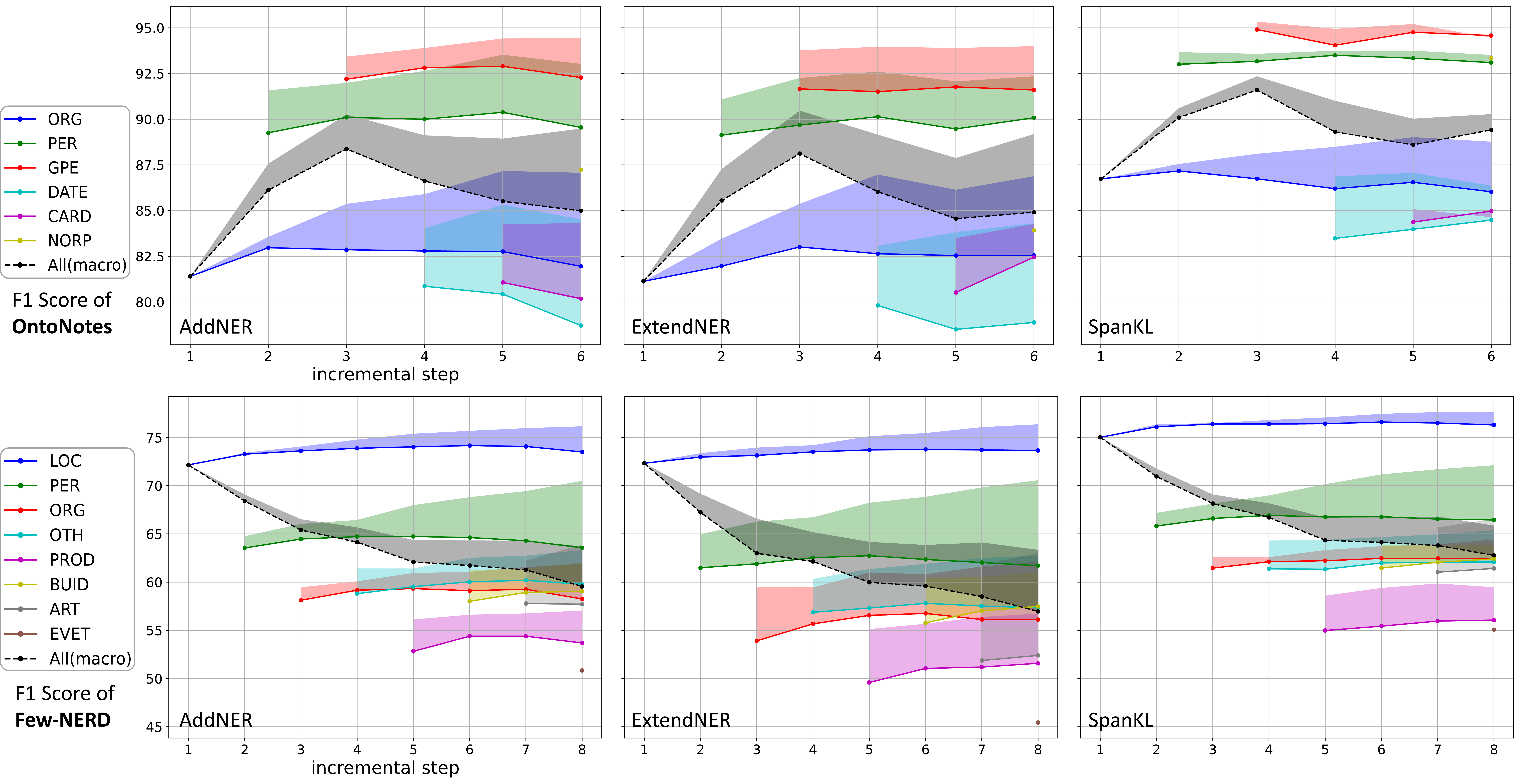}
\caption{Detailed F1 solid curve per entity type (different colors) at each step on OntoNotes/Few-NERD (rows) of different models (columns), with a certain learning order (shown downward on legend) in the Split-All setup.
Additional black dashed curves denote the macro-average over all types learned up to each step. Shaded areas of each curve denote the corresponding gap between CL and the upper non-CL.
}
\label{plot}
\end{figure*}

\begin{figure*}[htbp]
\centering
\includegraphics[width=6in]{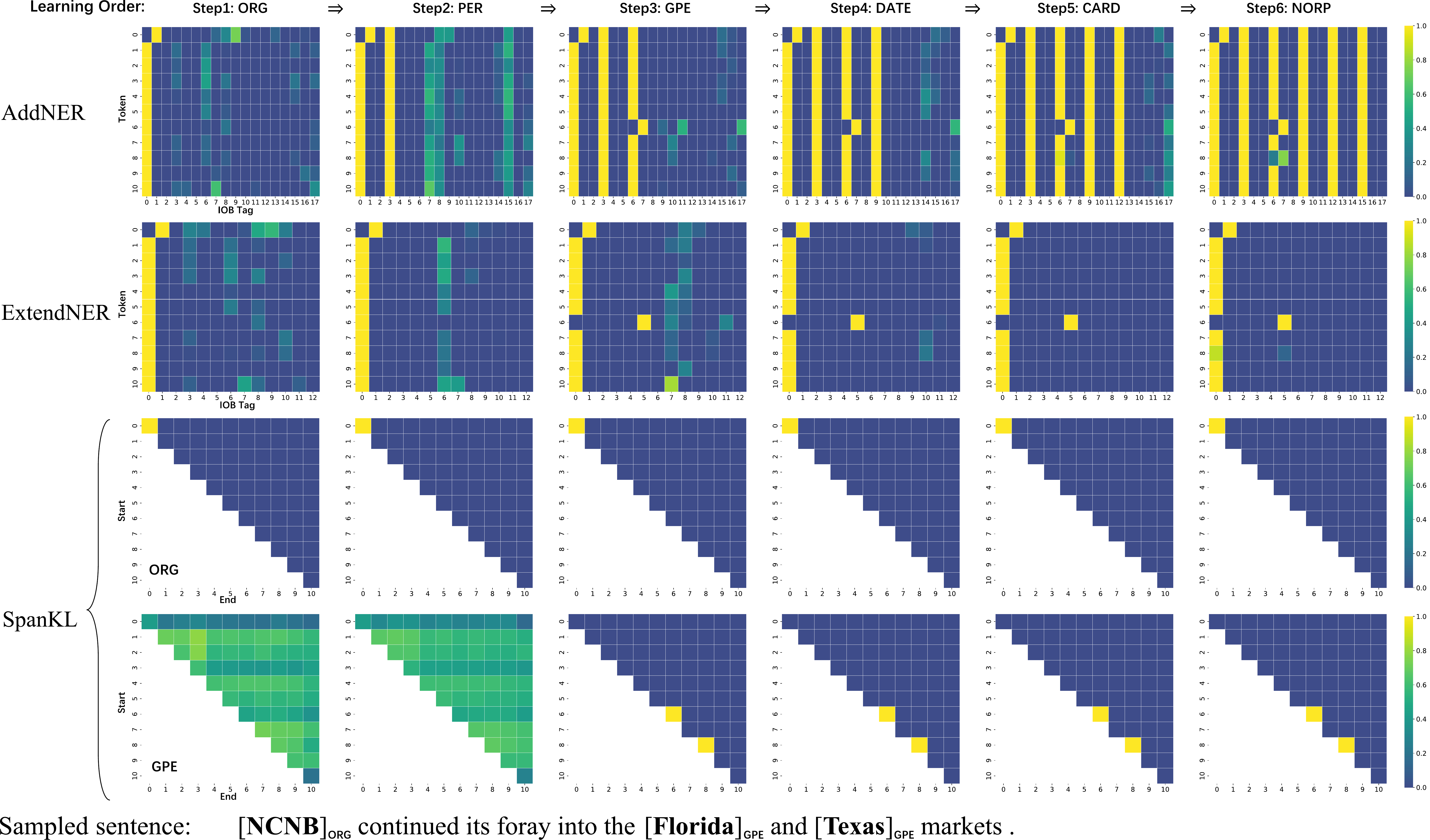}
\caption{Values probed from the normalized logit output per model at each step on OntoNotes (Split-All setup) with a certain learning order (shown in the upper). The rows and columns of every heatmap matrix in AddNER and ExtendNER denote the tokens and the predicted IOB tags. AddNER learns total $(1\!+\!2)\!\times\!6$ (\ttt{O} for each task) tags and ExtendNER learns total $1\!+\!2\!\times\!6$ (additional global \ttt{O}) tags at the final step. Span matrices in SpanKL relate to ORG and GPE. The random values in the place where the current task not yet to use are also shown.}
\label{probe}
\end{figure*}

\subsection{Results}
\noindent \tb{Synthetic Setups}. The Split-All, Split-Filter, Filter-All and Filter-Filter setups are preliminarily evaluated.
As shown in Tab.~\ref{performace_onto} on OntoNotes, the distinction between their performances for each model is consistent and also reasonable:
In terms of the adequacy of training data, Split-$\boldsymbol{\ast}$ setups contain plenty of non-entity samples whereas Filter-$\boldsymbol{\ast}$ do not, which promotes the training.
In terms of the difficulty of test data, $\boldsymbol{\ast}$-All are also more challenging than $\boldsymbol{\ast}$-Filter with the non-entity samples requiring the model's denial ability.
The overall performance ranking is: Split-Filter$>$Split-All$>$Filter-Filter$>$Filter-All, where Filter-All is the worst due to the contradictory requirements for the non-entity samples.

\noindent \tb{Overall Comparison}. On OntoNotes, SpanKL significantly outperforms others in three setups except for the irrational Filter-All setup.
It performs better at each step and is also closer to the upper bound all along.
Compared with the second best model accordingly, SpanKL achieves an absolute of 3.95, 3.78, 2.25 F1(\%) of improvements, and reduces the gaps by an absolute 3.48, 3.23, 0.83 into -0.76, -0.65, -3.83 at the final step, in Split-All, Split-Filter, Filter-Filter setups respectively.
Note that in these setups all models perform much the same in their non-CL settings but perform quite diversely in CL, revealing each model is theoretically strong enough to learn all task's entities but only stumbles in the CL settings.
In the most difficult Filter-All setup, SpanKL is slightly better than AddNER but with a little larger gap. And the ExtendNER is severely trapped in this setup.

Notably, we can verify the small performance gap between AddNER and ExtendNER indeed in Split-All as claimed in their paper. However, AddNER outperforms ExtendNER largely in typically Filter-$\x$ setups.
In addition to this finding, the recent SoTA L\&R, as claimed to be better than ExtendNER, is in fact worse than AddNER in their Filter-Filter setup. 
As expected, we confirm that AddNER is relatively better than ExtendNER with an \ttt{O} tag for each task (i,e. for each entity on OntoNotes) via sufficient test.
We also believe that SpanKL is further better by efficiently learning and transferring the information in the span and entity-level than the exhausted others that should combine multiple tags to represent entities.

We believe Split-All is the most common setup for CL typically without the redundant hypothesis. Therefore we mainly report results in this setup on the following Few-NERD evaluation (Tab.~\ref{performace_few}). Though the challenge that multiple entity types should be learned in each of the 8 tasks, SpanKL still substantial outperforms others, improving F1 over the second best AddNER at the final step by absolute 2.83, 2.76 and narrowing the gaps into -3.61, -3.63 in Split-All,
Split-Filter (listed deliberately by easily switching the test set),
respectively. We also find AddNER pulls away from ExtendNER in this complicated scenario as expected.

\noindent \tb{Entity-Level Comparison}. We explore the instant performance per entity at each step. We plot their F1 curves in different colors on both datasets with a certain learning order in Split-All.
We use an extra black dashed curve to denote the macro-average F1 over the learned entities and 
visualize the gap per entity using shaded areas, whose lower bound is the CL and the upper bound is the non-CL.

As shown in Fig.~\ref{plot}, although different entities have diverse performances due to their intrinsic difficulty, 
we still observe the SpanKL is more superior in that:
1) The starting point of every entity is consistently better than others, and almost keep this leading at each following step.
2) A flatter curve means less forgetting. Owing to KD all models have a flat curve in most of the entity types, but overall SpanKL is relatively more obvious.
3) Generally the shaded area of each entity type tends to be wider with the task increases, revealing the increasing hardship in CL approaching to the upper bound. But SpanKL still has much narrower shaded areas on both datasets especially the OntoNotes.

\noindent \tb{Probe of Model}.
For a deeper look, we probe the value (Fig.~\ref{probe}, explained in the caption) from the output on a sampled sentence from the test set of OntoNotes.
We observe that SpanKL successfully predicts all entities and doesn't forget, whereas AddNER and ExtendNER both fail in GPE entity \textit{Texas} even just after learning the GPE task.
All models achieve backward compatibility thanks to KD (i.e., keeping the same succeeding outputs once learned an entity).
AddNER, typically on OntoNotes, is also as forward compatible as SpanKL without requiring to change any preceding outputs. But SpanKL's learning capacity is still stronger possibly via span and entity-level separation as revealed by these multiple span matrices.
Yet AddNER is preferable to ExtendNER by an interesting finding on the auto-correction of the missing entity \textit{Texas} at its final step. We attribute this progressive generalization to the iterative self-supervised distillation. It also explains some growing curves in Fig.~\ref{plot} that the future distilled labels may be more accurate and consistent than the golden but possibly noisy labels initially learned.
Note that we do not perform the ablation study since SpanKL is very simple as designed without external architectural enhancers, and we hope it serves as a span-based immediate baseline in future works of CL-NER.  

\section{Conclusion}
In this paper, we propose a neural span-based model named SpanKL to serve as a strong baseline for CL-NER in class-incremental setting.
We empirically find that the independent modeling in span and entity-level is applicable to sequentially learning entity types, 
especially when cooperated with the KD technique and the multi-label prediction to easily attain coherent optimization. 
SpanKL closely approaches the upper bound of CL on OntoNotes with typically single entity per task, demonstrating its potential for practical application. It's still the best on the more complicated Few-NERD.
We also align and compare existing diverse synthetic setups for future research, whereby we validate AddNER is preferable to other sequence labeling models.

\section*{Acknowledgements}
We thank the reviewers for their thoughtful and constructive comments. This work was supported by the National Key R\&D Program of China (2021ZD0113402), the Science and Technology Planning Project of Shenzhen (JCYJ20190806112210067), the Natural Science Foundation of China (61872113, 62276075, 62276082), and the HIT(Shenzhen)-Tuling Robot Joint AL Lab.

\bibliography{aaai23}

\begin{thebibliography}{28}
\providecommand{\natexlab}[1]{#1}

\bibitem[{Biesialska, Biesialska, and
  Costa-juss{\`a}(2020)}]{biesialska-etal-2020-continual}
Biesialska, M.; Biesialska, K.; and Costa-juss{\`a}, M.~R. 2020.
\newblock Continual Lifelong Learning in Natural Language Processing: A Survey.
\newblock In \emph{Proceedings of the 28th International Conference on
  Computational Linguistics}, 6523--6541. Barcelona, Spain (Online):
  International Committee on Computational Linguistics.

\bibitem[{Castro et~al.(2018)Castro, Mar{\'\i}n-Jim{\'e}nez, Guil, Schmid, and
  Alahari}]{castro2018end}
Castro, F.~M.; Mar{\'\i}n-Jim{\'e}nez, M.~J.; Guil, N.; Schmid, C.; and
  Alahari, K. 2018.
\newblock End-to-end incremental learning.
\newblock In \emph{Proceedings of the European conference on computer vision
  (ECCV)}, 233--248.

\bibitem[{Chen and Moschitti(2019)}]{chen2019transfer}
Chen, L.; and Moschitti, A. 2019.
\newblock Transfer learning for sequence labeling using source model and target
  data.
\newblock In \emph{Proceedings of the AAAI Conference on Artificial
  Intelligence}, volume~33, 6260--6267.

\bibitem[{Chen and Liu(2018)}]{chen2018lifelong}
Chen, Z.; and Liu, B. 2018.
\newblock Lifelong machine learning.
\newblock \emph{Synthesis Lectures on Artificial Intelligence and Machine
  Learning}, 12(3): 1--207.

\bibitem[{De~Lange et~al.(2021)De~Lange, Aljundi, Masana, Parisot, Jia,
  Leonardis, Slabaugh, and Tuytelaars}]{de2021continual}
De~Lange, M.; Aljundi, R.; Masana, M.; Parisot, S.; Jia, X.; Leonardis, A.;
  Slabaugh, G.; and Tuytelaars, T. 2021.
\newblock A continual learning survey: Defying forgetting in classification
  tasks.
\newblock \emph{IEEE transactions on pattern analysis and machine
  intelligence}, 44(7): 3366--3385.

\bibitem[{Ding et~al.(2021)Ding, Xu, Chen, Wang, Han, Xie, Zheng, and
  Liu}]{ding2021few}
Ding, N.; Xu, G.; Chen, Y.; Wang, X.; Han, X.; Xie, P.; Zheng, H.-T.; and Liu,
  Z. 2021.
\newblock Few-NERD:A Few-shot Named Entity Recognition Dataset.
\newblock In \emph{ACL-IJCNLP}.

\bibitem[{Fu, Huang, and Liu(2021)}]{fu-2021-spanner}
Fu, J.; Huang, X.; and Liu, P. 2021.
\newblock {S}pan{NER}: Named Entity Re-/Recognition as Span Prediction.
\newblock In \emph{Proceedings of the 59th Annual Meeting of the Association
  for Computational Linguistics and the 11th International Joint Conference on
  Natural Language Processing (Volume 1: Long Papers)}, 7183--7195. Online:
  Association for Computational Linguistics.

\bibitem[{Goodfellow et~al.(2013)Goodfellow, Mirza, Xiao, Courville, and
  Bengio}]{goodfellow2013empirical}
Goodfellow, I.~J.; Mirza, M.; Xiao, D.; Courville, A.; and Bengio, Y. 2013.
\newblock An empirical investigation of catastrophic forgetting in
  gradient-based neural networks.
\newblock \emph{arXiv preprint arXiv:1312.6211}.

\bibitem[{Gupta et~al.(2020)Gupta, Chaudhary, Runkler, and
  Schütze}]{DBLP:conf/icml/GuptaCRS20}
Gupta, P.; Chaudhary, Y.; Runkler, T.~A.; and Schütze, H. 2020.
\newblock Neural Topic Modeling with Continual Lifelong Learning.
\newblock In \emph{ICML}, 3907--3917.

\bibitem[{Hinton et~al.(2015)Hinton, Vinyals, Dean
  et~al.}]{hinton2015distilling}
Hinton, G.; Vinyals, O.; Dean, J.; et~al. 2015.
\newblock Distilling the knowledge in a neural network.
\newblock \emph{arXiv preprint arXiv:1503.02531}, 2(7).

\bibitem[{Kirkpatrick et~al.(2017)Kirkpatrick, Pascanu, Rabinowitz, Veness,
  Desjardins, Rusu, Milan, Quan, Ramalho, Grabska-Barwinska
  et~al.}]{kirkpatrick2017overcoming}
Kirkpatrick, J.; Pascanu, R.; Rabinowitz, N.; Veness, J.; Desjardins, G.; Rusu,
  A.~A.; Milan, K.; Quan, J.; Ramalho, T.; Grabska-Barwinska, A.; et~al. 2017.
\newblock Overcoming catastrophic forgetting in neural networks.
\newblock \emph{Proceedings of the national academy of sciences}, 114(13):
  3521--3526.

\bibitem[{Lample et~al.(2016)Lample, Ballesteros, Subramanian, Kawakami, and
  Dyer}]{lample-2016-lstmcrfner}
Lample, G.; Ballesteros, M.; Subramanian, S.; Kawakami, K.; and Dyer, C. 2016.
\newblock Neural Architectures for Named Entity Recognition.
\newblock In \emph{Proceedings of the 2016 Conference of the North {A}merican
  Chapter of the Association for Computational Linguistics: Human Language
  Technologies}, 260--270. San Diego, California: Association for Computational
  Linguistics.

\bibitem[{Li and Hoiem(2017)}]{li2017learning}
Li, Z.; and Hoiem, D. 2017.
\newblock Learning without forgetting.
\newblock \emph{IEEE transactions on pattern analysis and machine
  intelligence}, 40(12): 2935--2947.

\bibitem[{Loshchilov and Hutter(2017)}]{loshchilov2017decoupled}
Loshchilov, I.; and Hutter, F. 2017.
\newblock Decoupled weight decay regularization.
\newblock \emph{arXiv preprint arXiv:1711.05101}.

\bibitem[{Masana et~al.(2020)Masana, Liu, Twardowski, Menta, Bagdanov, and
  van~de Weijer}]{masana2020class}
Masana, M.; Liu, X.; Twardowski, B.; Menta, M.; Bagdanov, A.~D.; and van~de
  Weijer, J. 2020.
\newblock Class-incremental learning: survey and performance evaluation on
  image classification.
\newblock \emph{arXiv preprint arXiv:2010.15277}.

\bibitem[{Monaikul et~al.(2021)Monaikul, Castellucci, Filice, and
  Rokhlenko}]{Monaikul2021ContinualLF}
Monaikul, N.; Castellucci, G.; Filice, S.; and Rokhlenko, O. 2021.
\newblock Continual Learning for Named Entity Recognition.
\newblock In \emph{AAAI}.

\bibitem[{Pradhan et~al.(2013)Pradhan, Elhadad, South, Martinez, Vogel,
  Suominen, Chapman, and Savova}]{Pradhan13ontonote}
Pradhan, S.; Elhadad, N.; South, B.~R.; Martinez, D.; Vogel, A.; Suominen, H.;
  Chapman, W.~W.; and Savova, G. 2013.
\newblock Task 1: ShARe/CLEF eHealth Evaluation Lab.

\bibitem[{Rebuffi et~al.(2017)Rebuffi, Kolesnikov, Sperl, and
  Lampert}]{rebuffi2017icarl}
Rebuffi, S.-A.; Kolesnikov, A.; Sperl, G.; and Lampert, C.~H. 2017.
\newblock icarl: Incremental classifier and representation learning.
\newblock In \emph{Proceedings of the IEEE conference on Computer Vision and
  Pattern Recognition}, 2001--2010.

\bibitem[{Serra et~al.(2018)Serra, Suris, Miron, and
  Karatzoglou}]{serra2018overcoming}
Serra, J.; Suris, D.; Miron, M.; and Karatzoglou, A. 2018.
\newblock Overcoming catastrophic forgetting with hard attention to the task.
\newblock In \emph{International Conference on Machine Learning}, 4548--4557.
  PMLR.

\bibitem[{Shin et~al.(2017)Shin, Lee, Kim, and Kim}]{shin2017continual}
Shin, H.; Lee, J.~K.; Kim, J.; and Kim, J. 2017.
\newblock Continual learning with deep generative replay.
\newblock \emph{Advances in neural information processing systems}, 30.

\bibitem[{Su(2021)}]{kexuefm-8373}
Su, J. 2021.
\newblock GlobalPointer: A unified way to handle nested and flat named entity
  recognition.
\newblock \url{https://kexue.fm/archives/8373}.

\bibitem[{Van~de Ven and Tolias(2019)}]{van2019three}
Van~de Ven, G.~M.; and Tolias, A.~S. 2019.
\newblock Three scenarios for continual learning.
\newblock \emph{arXiv preprint arXiv:1904.07734}.

\bibitem[{Wang et~al.(2021)Wang, Jiang, Yan, Jia, Bach, Wang, Huang, Huang, and
  Tu}]{wang-etal-2021-structural}
Wang, X.; Jiang, Y.; Yan, Z.; Jia, Z.; Bach, N.; Wang, T.; Huang, Z.; Huang,
  F.; and Tu, K. 2021.
\newblock Structural Knowledge Distillation: Tractably Distilling Information
  for Structured Predictor.
\newblock In \emph{Proceedings of the 59th Annual Meeting of the Association
  for Computational Linguistics and the 11th International Joint Conference on
  Natural Language Processing (Volume 1: Long Papers)}, 550--564. Online:
  Association for Computational Linguistics.

\bibitem[{Wolf et~al.(2019)Wolf, Debut, Sanh, Chaumond, Delangue, Moi, Cistac,
  Rault, Louf, Funtowicz et~al.}]{wolf2019huggingface}
Wolf, T.; Debut, L.; Sanh, V.; Chaumond, J.; Delangue, C.; Moi, A.; Cistac, P.;
  Rault, T.; Louf, R.; Funtowicz, M.; et~al. 2019.
\newblock HuggingFace's Transformers: State-of-the-art Natural Language
  Processing.
\newblock \emph{ArXiv}, arXiv--1910.

\bibitem[{Xia et~al.(2022)Xia, Wang, Lyu, Zhu, Wu, Li, and
  Dai}]{xia-etal-2022-learn}
Xia, Y.; Wang, Q.; Lyu, Y.; Zhu, Y.; Wu, W.; Li, S.; and Dai, D. 2022.
\newblock Learn and Review: Enhancing Continual Named Entity Recognition via
  Reviewing Synthetic Samples.
\newblock In \emph{Findings of the Association for Computational Linguistics:
  ACL 2022}, 2291--2300. Dublin, Ireland: Association for Computational
  Linguistics.

\bibitem[{Xu and Zhu(2018)}]{xu2018reinforced}
Xu, J.; and Zhu, Z. 2018.
\newblock Reinforced continual learning.
\newblock \emph{Advances in Neural Information Processing Systems}, 31.

\bibitem[{Xu et~al.(2021)Xu, Huang, Feng, and Hu}]{Xu-2021-SMHSA}
Xu, Y.; Huang, H.; Feng, C.; and Hu, Y. 2021.
\newblock A Supervised Multi-Head Self-Attention Network for Nested Named
  Entity Recognition.
\newblock In \emph{Thirty-Fifth {AAAI} Conference on Artificial Intelligence,
  {AAAI}}, 14185--14193. {AAAI} Press.

\bibitem[{Yu, Bohnet, and Poesio(2020)}]{yu-2020-biaffine:span-based}
Yu, J.; Bohnet, B.; and Poesio, M. 2020.
\newblock Named Entity Recognition as Dependency Parsing.
\newblock In \emph{Proceedings of the 58th Annual Meeting of the Association
  for Computational Linguistics}, 6470--6476. Online: Association for
  Computational Linguistics.

\end{thebibliography}

\section*{Appendix}
\subsection{Permutations of Tasks}
As described, we follow the existing 6 permutations of tasks on OntoNotes from recent works, and randomly sample 4 permutations on Few-NERD to factor out the influence of the learning order as shown in Tab.~\ref{perm_order}.

\begin{table}[htbp]\small
\tabcolsep=2pt
\setlength{\belowcaptionskip}{-0.2cm}
\begin{tabular}{llllllll}
\toprule
\multicolumn{8}{c}{6 OntoNotes Permutations ($\Longrightarrow$)}             \\
\midrule
1: ORG  & PER  & GPE   & DATE & CARD & NORP &      &      \\ \hline
2: DATE & NORP & PER   & CARD & ORG  & GPE  &      &      \\ \hline
3: GPE  & CARD & ORG   & NORP & DATE & PER  &      &      \\ \hline
4: NORP & ORG  & DATE  & PER  & GPE  & CARD &      &      \\ \hline
5: CARD & GPE  & NORP  & ORG  & PER  & DATE &      &      \\ \hline
6: PER  & DATE & CARD  & GPE  & NORP & ORG  &      &      \\
\bottomrule
\toprule
\multicolumn{8}{c}{4 Few-NERD Permutations ($\Longrightarrow$)}              \\
\midrule
1: LOC  & PER  & ORG   & OTH  & PROD & BUID & ART  & EVET \\ \hline
2: ORG  & PROD & ART   & EVET & OTH  & PER  & LOC  & BUID \\ \hline
3: PROD & EVET & OTH   & PER  & ART  & LOC  & BUID & ORG  \\ \hline
4: BUID & OTH  & PROD  & PER  & ORG  & LOC  & ART  & EVET \\

\bottomrule
\end{tabular}
\caption{Different permutations of tasks used on OntoNotes and Few-NERD.}
\label{perm_order}
\end{table}

\subsection{Fine-grained Entity Types on Few-NERD}
All the fine-grained entity types on Few-NERD will be used in our evaluation, as shown in Tab.~\ref{finegraind}.
The corresponding coarse-grained entity types are bold.

\begin{table}[!h]
\center
\begin{tabular}{|l|l|}
\hline
\multirow{7}{*}{\textbf{location}}      & bodiesofwater                      \\
                                        & GPE                                \\
                                        & island                             \\
                                        & mountain                           \\
                                        & other                              \\
                                        & park                               \\
                                        & road/railway/highway/transit       \\ \hline
\multirow{8}{*}{\textbf{person}}        & actor                              \\
                                        & artist/author                      \\
                                        & athlete                            \\
                                        & director                           \\
                                        & other                              \\
                                        & politician                         \\
                                        & scholar                            \\
                                        & soldier                            \\ \hline

\end{tabular}
\end{table}

\begin{table}[!h]
\center
\begin{tabular}{|l|l|}
\hline

\multirow{10}{*}{\textbf{organization}} & company                            \\
                                        & education                          \\
                                        & government/governmentagency        \\
                                        & media/newspaper                    \\
                                        & other                              \\
                                        & politicalparty                     \\
                                        & religion                           \\
                                        & showorganization                   \\
                                        & sportsleague                       \\
                                        & sportsteam                         \\ \hline
\multirow{12}{*}{\textbf{other}}        & astronomything                     \\
                                        & award                              \\
                                        & biologything                       \\
                                        & chemicalthing                      \\
                                        & currency                           \\
                                        & disease                            \\
                                        & educationaldegree                  \\
                                        & god                                \\
                                        & language                           \\
                                        & law                                \\
                                        & livingthing                        \\
                                        & medical                            \\ \hline
\multirow{9}{*}{\textbf{product}}       & airplane                           \\
                                        & car                                \\
                                        & food                               \\
                                        & game                               \\
                                        & other                              \\
                                        & ship                               \\
                                        & software                           \\
                                        & train                              \\
                                        & weapon                             \\ \hline
\multirow{8}{*}{\textbf{build}}         & airport                            \\
                                        & hospital                           \\
                                        & hotel                              \\
                                        & library                            \\
                                        & other                              \\
                                        & restaurant                         \\
                                        & sportsfacility                     \\
                                        & theater                            \\ \hline
\multirow{6}{*}{\textbf{art}}           & broadcastprogram                   \\
                                        & film                               \\
                                        & music                              \\
                                        & other                              \\
                                        & painting                           \\
                                        & writtenart                         \\ \hline
\multirow{6}{*}{\textbf{event}}         & attack/battle/war/militaryconflict \\
                                        & disaster                           \\
                                        & election                           \\
                                        & other                              \\
                                        & protest                            \\
                                        & sports                             \\ \hline
\end{tabular}
\caption{The fine-grained entity types on Few-NERD}
\label{finegraind}
\end{table}

\subsection{Performance w.r.t Different Permutations}
To investigate the impact on the final performance of whether an entity type is learned earlier or later,
we further plot the performances regarding all permutations on both datasets in Split-All setup of SpanKL.
Similarly, we also use an extra black dashed curve to denote the Macro-F1 score over all entity types learned so far.

\begin{figure*}[htbp]
\centering
\includegraphics[width=6in]{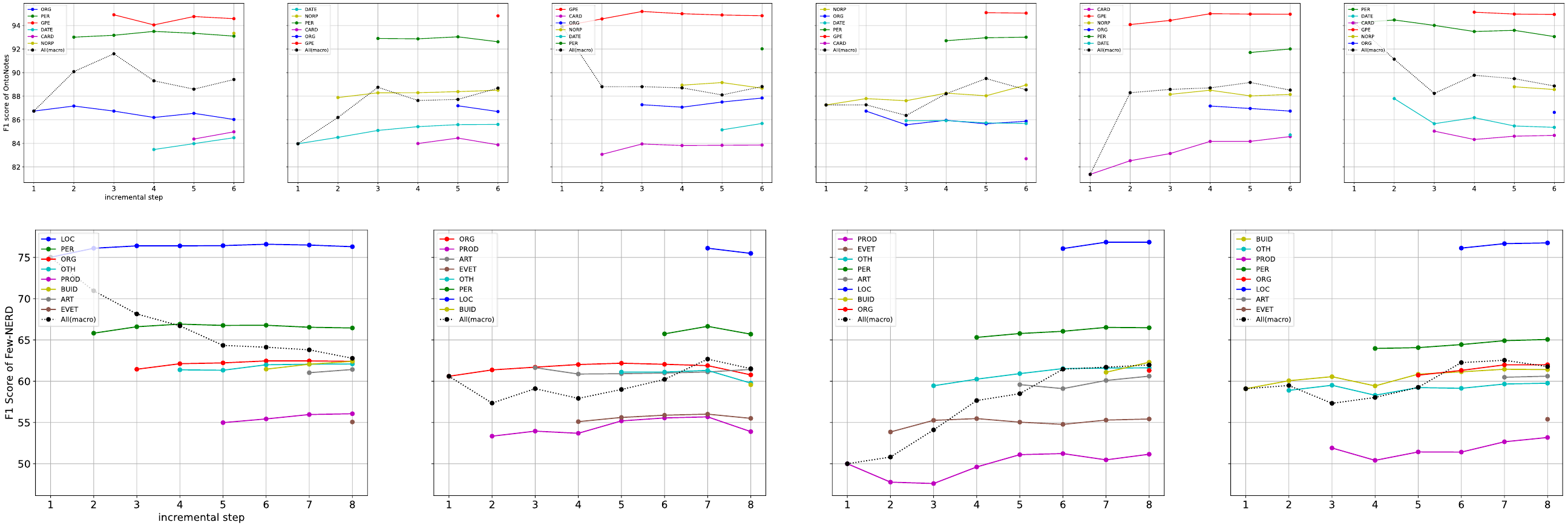}
\caption{F1 curves of different entity types on OnteNotes and Few-NERD (rows) with different learning order (columns) indicated by downward in legends.
}
\label{perm_evaluate}
\end{figure*}

As shown in Fig.~\ref{perm_evaluate}, we can observe that most entity types have a similar performance eventually although start to be learned at different steps.
The final Macro-F1 scores in different orders also converge to a stable point. We can presume that the result at the final step of a certain permutation is already relatively representative for evaluation on CL.

\end{document}